\documentclass[a4]{article}

\usepackage[margin=1in]{geometry}
\usepackage{authblk}

\newif\ifdraft %
\draftfalse

\usepackage{graphicx}
\usepackage{amsmath}
\usepackage{amssymb}
\usepackage{color}
\usepackage{here}
\usepackage{url}
\usepackage{subfig}

\usepackage{bm}

\usepackage[ruled,linesnumbered,vlined]{algorithm2e}
\SetKwBlock{Function}{function}{end function}
\SetKwComment{CommentTmp}{\blue{$\triangleright$\:}}{}
\DontPrintSemicolon
\newcommand{\Comment}[1]{\CommentTmp*[r]{\small\textcolor{blue}{#1}\!\!\!\!\!\!}}

\ifdraft
\newcommand{\red}[1]{\textcolor{red}{#1}}

\newcommand{\blue}[1]{\textcolor{blue}{#1}}
\else
\newcommand{\red}[1]{#1}

\newcommand{\blue}[1]{}
\fi

\def\*#1{\boldsymbol{#1}}

\newtheorem{theo}{Theorem}[section]

\newcommand{\tw}[0]{\textwidth}
\newcommand{\igr}[2]{\includegraphics[clip,width=#1\tw]{#2}}

\newcommand{\pd}[2]{{\frac{\partial #1}{\partial #2}}}

\newcommand{\eq}[1]{(\ref{#1})}

\newcommand{\lw}[1]{\smash{\lower2.ex\hbox{#1}}}

\newcommand{\RR}{\mathbb{R}}

\newcommand{\cC}{{\mathcal C}}

\newcommand{\cH}{{\mathcal H}}
\newcommand{\cI}{{\mathcal I}}

\newcommand{\cL}{{\mathcal L}}

\newcommand{\cW}{{\mathcal W}}

\newcommand{\cZ}{{\mathcal Z}}

\title{
Learning Attributed Graphlets: \\
Predictive Graph Mining by Graphlets with Trainable Attribute
}

\author[1]{Tajima Shinji}
\author[1]{Ren Sugihara}
\author[1]{Ryota Kitahara}
\author[1]{Masayuki Karasuyama$^*$}

\affil[1]{Nagoya Institute of Technology}
\affil[$*$]{\protect\url{karasuyama@nitech.ac.jp}}

\date{}

\begin{document}

\maketitle

\begin{abstract}
The graph classification problem has been widely studied; however, achieving an interpretable model with high predictive performance remains a challenging issue. 
This paper proposes an interpretable classification algorithm for attributed graph data, called LAGRA (Learning Attributed GRAphlets). 
LAGRA learns importance weights for small attributed subgraphs, called attributed graphlets (AGs), while simultaneously optimizing their attribute vectors. 
This enables us to obtain a combination of subgraph structures and their attribute vectors that strongly contribute to discriminating different classes. 
A significant characteristics of LAGRA is that all the subgraph structures in the training dataset can be considered as a candidate structures of AGs. 
This approach can explore all the potentially important subgraphs exhaustively, but obviously, a na{\"i}ve implementation can require a large amount of computations. 
To mitigate this issue, we propose an efficient pruning strategy by combining the proximal gradient descent and a graph mining tree search. 
Our pruning strategy can ensure that the quality of the solution is maintained compared to the result without pruning. 
We empirically demonstrate that LAGRA has superior or comparable prediction performance to the standard existing algorithms including graph neural networks, while using only a small number of AGs in an interpretable manner. 
\end{abstract}

\section{Introduction}
\label{sec:intro}

Prediction problems with a graph input, such as graph classification problems, have been widely studied in the data science community.
A graph representation is useful to capture structural data, and graph-based machine learning algorithms have been applied to variety of application problems such as chemical composition analysis \cite{ralaivola2005graph,faber2017prediction} and crystal structure analysis \cite{xie2018crystal,louis2020graph}.
In real-word datasets, graphs often have node attributes as a continuous value vector (note that we only focus on node attributes throughout the paper, but the discussion is same for edge attributes).
%
For example, a graph created by a chemical composition can have a three dimensional position of each atom as an attribute vector in addition to a categorical label such as atomic species. 
In this paper, we consider building an {\it interepretable} prediction model for a graph classification problem in which an input graph has continuous attribute vectors. 
As we will see in Section~\ref{sec:related-work}, this setting has not been widely studied despite its practical importance.

\begin{figure}
 \begin{center}
  \igr{.6}{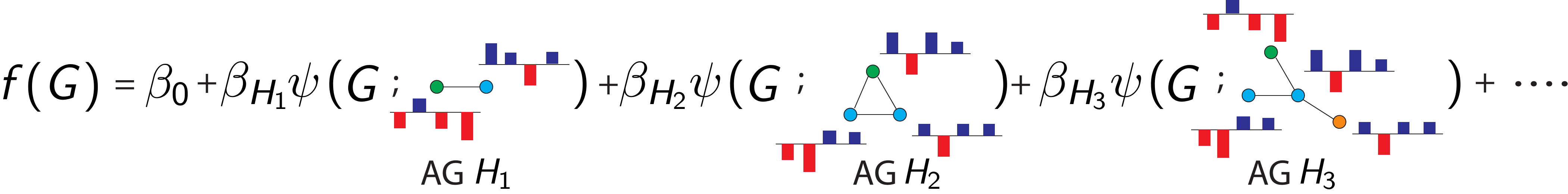}
 \end{center}
 \caption{
 Illustration of our attributed graphlet (AG) based prediction model.
 %
 %
 The colors of each graph node represents a graph node label, and a bar plot associated with each graph node represents a trainable attribute vector.
 }
 \label{fig:model}
\end{figure}

Our framework can identify important small subgraphs, called graphlets, in which each node has an attribute vector.
Note that we use the term graphlet simply to denote a small connected subgraph \cite{shervashidze2009efficient}, 
though in some papers, it only indicates induced subgraphs \cite{przulj2007biological}. 
%
%
\figurename~\ref{fig:model} shows an illustration of our prediction model. 
In the figure, the output of the prediction model
$f(G) = \beta_0 + \beta_{H_1} \psi(G ; H_1) + \beta_{H_2} \psi(G ; H_2)) + \cdots$ 
for an input attributed graph $G$ is defined through a linear combination of {\it attributed graphlets} (AGs), represented as $H_1, H_2, \ldots$, each one of which is weighted by parameters $\beta_{H_1}, \beta_{H_2}, \ldots$.
The function $\psi(G ; H)$ evaluates a matching score between $G$ and an AG $H$ in a sense that how precisely $G$ contains the AG $H$. 
%
%
We apply a sparse regularization to the parameter $\beta_{H}$ by which a small number of important AGs for classification can be obtained, i.e., an AG with non-zero $\beta_{H}$ (in particular, if it has large $|\beta_{H}|$) can be regarded as a discriminative AG.

\begin{figure}
 \begin{center}
  \igr{.6}{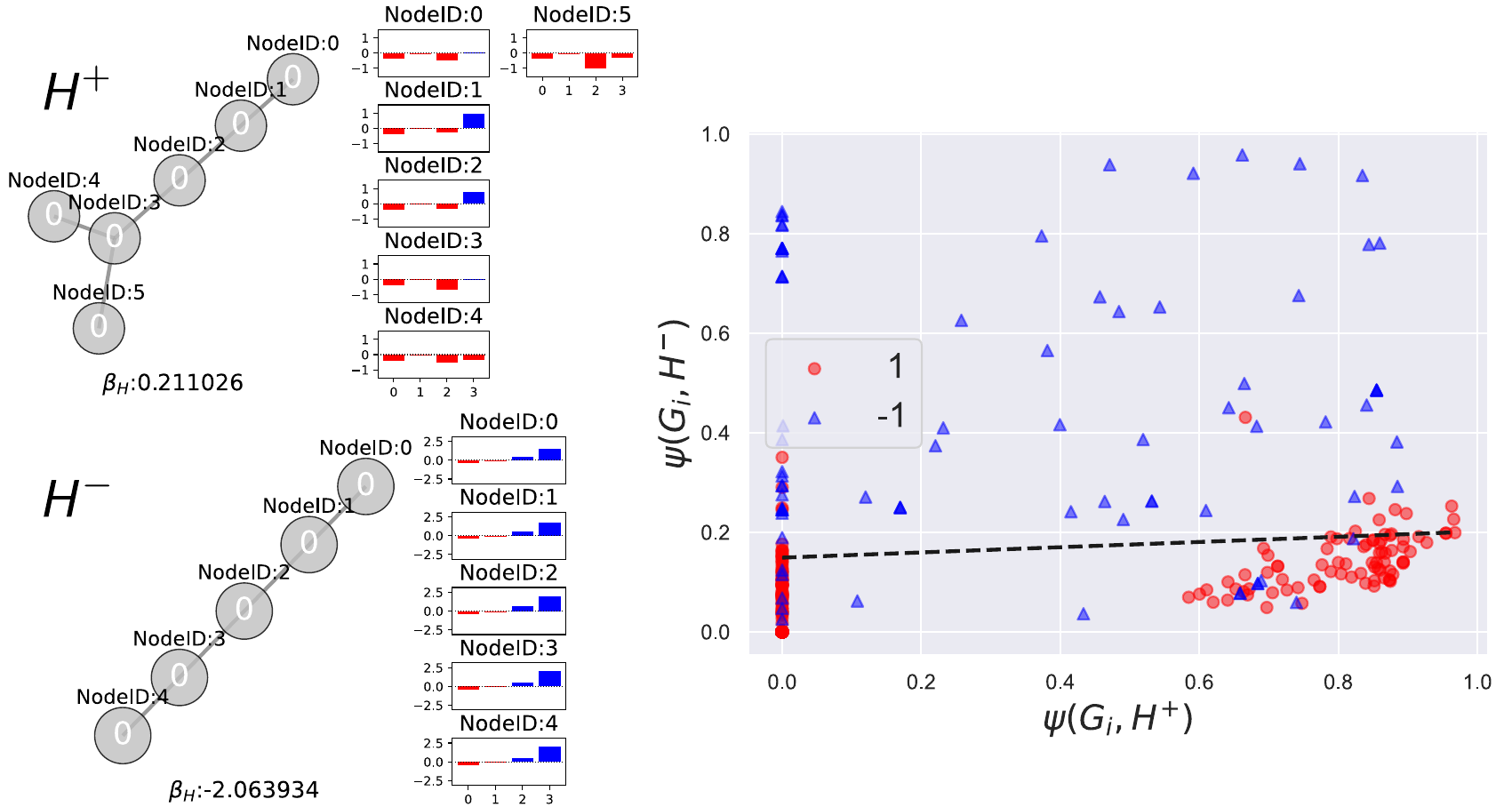}
 \end{center}
 \caption{\red{
 An example of important attributed graphlets $H^+$ and $H^-$ identified by LAGRA in the AIDS dataset.
 $H^+$ and $H^-$ positively and negatively contribute to the prediction, respectively.
 The right plot is a scatter in which $x$- and $y$- axes are our graphlet features representing the how precisely $H^+$ and $H^-$ are included in the input graph $G_i$.
 Each point is from the test dataset. 
 }}
 \label{fig:scatter-AIDS}
\end{figure}

Important subgraphs and attribute vectors are usually unknown beforehand.
The basic strategy of our proposed method, called LAGRA (Learning Attributed GRAphlets), is as follows:
\begin{itemize}
 \item To explore potentially important substructures of graphs, i.e., subgraphs, LAGRA uses {\it graph mining} by which all the subgraphs in the given dataset can be considered up to the given maximum graph size. 

 \item For continuous attributes in AGs, we optimize them as trainable parameters.
\end{itemize}
%
%
\red{
\figurename~\ref{fig:scatter-AIDS} is an example of identified AGs by LAGRA, which shows only two AGs clearly separate two classes (See Section~\ref{ssec:examples-selected-AG} for detail).
}
%
Since the number of the possible subgraphs is quite large and an attribute vector exists for each node in each one of subgraphs, a na{\"i}ve implementation becomes computationally intractable.
For the efficient optimization, we employ a block coordinate update \cite{xu2017globally} based approach in which $\*\beta$ (a vector containing $\beta_{H}$), the bias term $\beta_0$, and attribute vectors are alternately updated.
In the alternate update of $\*\beta$, we apply the proximal gradient descent \cite{Teboulle2017simplified,bech2009fast}, which is known as an effective algorithm to optimize sparse parameters. 
For this step, we propose an efficient pruning strategy, enabling us to identify dimensions that are not required to update at that iteration. 
This pruning strategy has the three advantages.
First, by combining the sparsity during the proximal update and the graph mining tree search, we can eliminate unnecessary dimensions without enumerating all the possible subgraphs.
Second, for removed variables $\beta_{H}$ at that iteration, attribute vectors in $H$ are also not required to be updated, which also accelerates the optimization.
Third, our pruning strategy is designed so that it can maintain the update result compared with when we do not perform the pruning (In other words, our pruning strategy does not deteriorate the resulting model accuracy).

Our contributions are summarized as follows:
\begin{itemize}
 \item We propose an interpretable graph classification model, in which the prediction is defined through a linear combination of graphlets that have trainable attribute vectors.
       By imposing a sparse penalty on the coefficient of each AG, a small number of important AGs can be identified. 

 \item To avoid directly handling an intractably large size of optimization variables, we propose an efficient pruning strategy \red{based on the proximal gradient descent}, which can safely ignore AGs that do not contribute to the update.

 \item We verify effectiveness of LAGRA by empirical evaluations. 
       Although our prediction model is simple and interpretable, we show that prediction performance of LAGRA was superior to or comparable with well-known standard graph classification methods, and in those results, LAGRA actually only used a small number of AGs.
       Further, we also show examples of selected AGs to demonstrate the high interpretability.
      
\end{itemize}

\section{Proposed Method: LAGRA}
\label{sec:proposed-method}

In this section, we describe our proposed method, called Learning Attributed GRAphlets (LAGRA).
First, in Section~\ref{ssec:formulation}, we show the formulation of our model and the definition of the optimization problem.
Second, in Section~\ref{ssec:optimization}, we show an efficient optimization algorithm for LAGRA.

\subsection{Formulation}
\label{ssec:formulation}

\subsubsection{Problem Setting}
\label{sssec:problem-setting}

We consider a classification problem in which a graph $G$ is an input. 
A set of nodes and edges of $G$ are written as $V_G$ and $E_G$, respectively.
Each one of nodes $v \in V_G$ has a categorical label $L_v$ and a continuous attribute vector $\*z^G_v \in \RR^d$, where $d$ is an attribute dimension. 
In this paper, an attribute indicates a continuous attribute vector. 
We assume that a label and an attribute vector are for a node, but the discussion in this paper is completely same as for an edge label and attribute.
A training dataset is 
$\{ (G_i, y_i) \}_{i \in [n]}$,
in which $y_i \in \{ -1, +1 \}$ is a binary label and $n$ is the dataset size, where $[n] = \{ 1, \ldots, n\}$.
Although we only focus on the classification problem, our framework is also applicable to the regression problem just by replacing the loss function.

\subsubsection{Attributed Graphlet Inclusion Score}
\label{sssec:AGIS}

We consider extracting important small attributed graphs, which we call {\it attributed graphlets} (AGs), that contributes to the classification boundary. 
Note that throughout the paper, we only consider a connected graph as an AG for a better interpretability (do not consider an AG by a disconnected graph).
Let 
$\psi(G_i; H) \in [0,1]$
be a feature representing a degree that an input graph includes an AG $H$. 
We refer to $\psi(G_i; H)$ as the {\it AG inclusion score} (AGIS).
Our proposed LAGRA identifies important AGs by applying a feature selection to a model with this AGIS feature.

\begin{figure}
 \begin{center}
  \igr{.4}{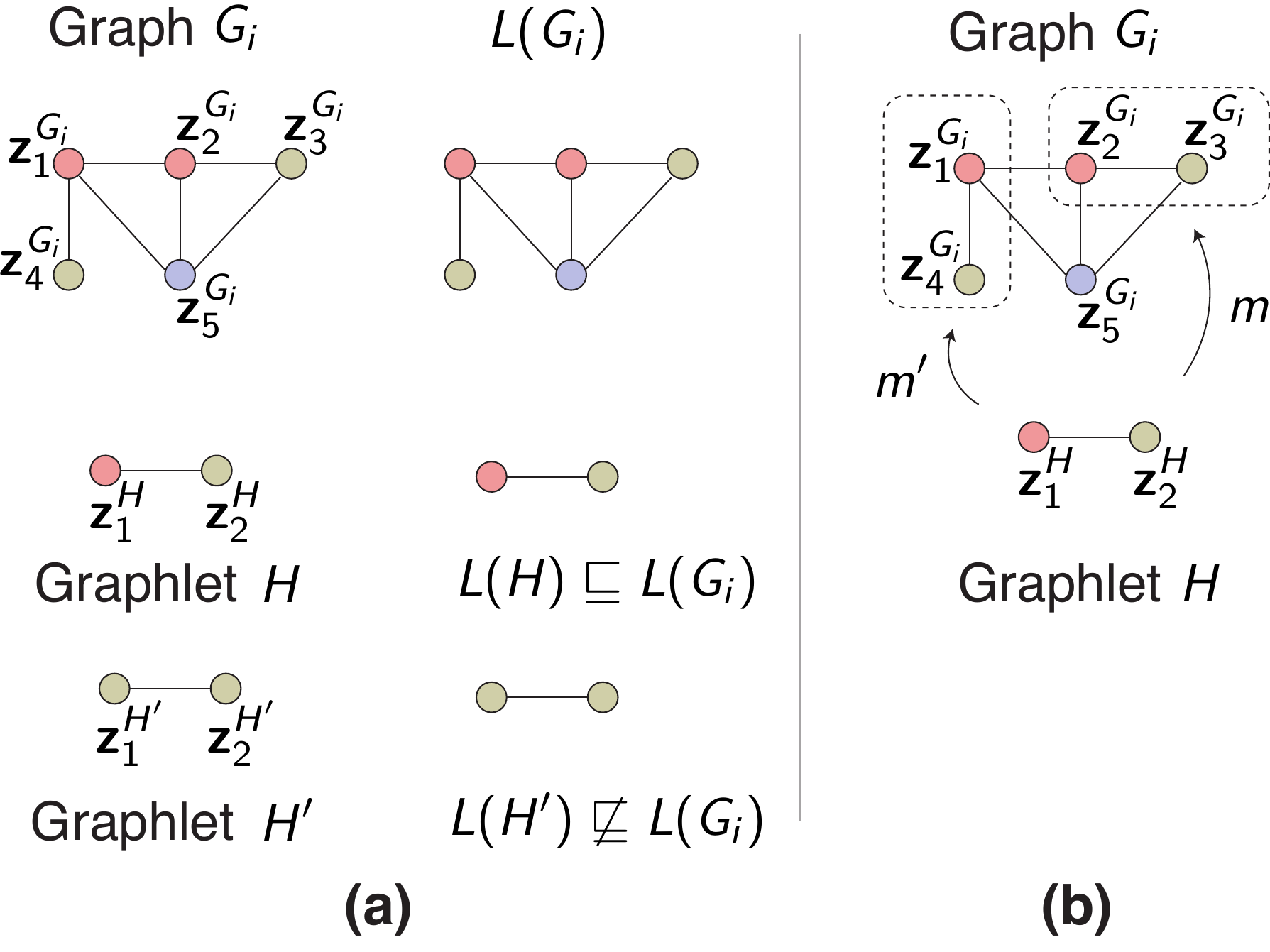}
 \end{center}
 \caption{
 Examples of matchings between a graph and AGs (colors of graph nodes are node labels).
 (a) For two AGs $H$ and $H^\prime$, $L(G_i)$ only contains $L(H)$, and $L(H^\prime)$ is not contained.
 Then, $\psi(G_i ; H) > 0$ and $\psi(G_i ; H^\prime) = 0$.
 (b)
 An example of the set of injections 
 $M = \{ m, m^\prime \}$, 
 where 
 $m(1) = 2, m(2) = 3, m^\prime(1) = 1$, 
 and 
 $m^\prime(2) = 4$.
 The figure shows that
 $m$ and $m^\prime$ are label and edge preserving.
 }
 \label{fig:graph-matching}
\end{figure}

Suppose that $L(G)$ is a labeled graph having a categorical label $L_v$ for each node, and in $L(G)$, an attribute $\*z^G_v$ for each node is excluded from $G$.
We define AGIS so that it has a non-zero value only when $L(H)$ is included in $L(G_i)$:
%
\begin{align}
 \psi(G_i ; H) = 
 \begin{cases}
  \phi_H(G_i) & \text{if } L(H) \sqsubseteq L(G_i), \\
  0 & \text{otherwise}, 
 \end{cases}
 \label{eq:AGIS}
\end{align}
where 
$L(H) \sqsubseteq L(G_i)$ 
means that $L(H)$ is a subgraph of $L(G_i)$, and 
$\phi_H(G_i) \in (0,1]$
is a function that provides a continuous inclusion score of $H$ in $G_i$.   
The condition $L(H) \sqsubseteq L(G_i)$ makes AGIS highly interpretable.
For example, in the case of chemical composition data, if $L(H)$ represents O-C (oxygen and carbon are connected) and $\psi(G_i ; H) > 0$, then we can guarantee that $G_i$ must contain O-C.
%
\figurename~\ref{fig:graph-matching}(a) shows examples of $L(G_i)$ and $L(H)$.

The function
$\phi_H(G_i)$
needs to be defined so that it can represent how strongly the attribute vectors in $H$ can be matched to those of $G_i$.
When $L(H) \sqsubseteq L(G_i)$, there exists at least one injection 
$m: V_H \rightarrow V_{G_i}$
in which $m(v)$ for $v \in V_H$ preserves node labels and edges among $v \in V_H$.
\figurename~\ref{fig:graph-matching}(b) shows an example of when there exist two injections.
Let $M$ be a set of possible injections $m$. 
%
\red{
We define a similarity between $H$ and a subgraph of $G_i$ matched by $m \in M$ as follows 
\begin{align*}
 \mathrm{Sim}(H, G_i ; m) = 
 \exp\left( - \rho 
 \sum_{v \in V_H} \left\| \*z^{H}_v - \*z^{G_i}_{m(v)} \right\|^2
 \right),
\end{align*}
where $\rho > 0$ is a fixed parameter that adjusts the length scale. 
In $\exp$, the sum of squared distances of attribute vectors between matched nodes are taken.
To use this similarity in AGIS \eqref{eq:AGIS}, we take the maximum among all the matchings $M$:
\begin{align}
 \phi_H(G_i) = \mathrm{MaxPooling}(\left\{ \mathrm{Sim}(H, G_i ; m ) : m \in M \right\})
 \label{eq:phi}
\end{align}
}
%
%
An intuition behind \eq{eq:phi} is that it evaluates inclusion of $H$ in $G_i$ based on the best macthing in a sense of 
$\mathrm{Sim}(H, G_i ; m)$
for $m \in M$.
%
If $L(H) \sqsubseteq L(G_i)$ and there exists $m$ such that $\*z^H_{v} = \*z^{G_i}_{m(v)}$ for $\forall v \in V_H$, then, $\phi_H(G_i)$ takes the maximum value (i.e., $1$).

\subsubsection{Model definition}
\label{sssec:model-definition}

\begin{figure}
 \begin{center}
  \igr{.35}{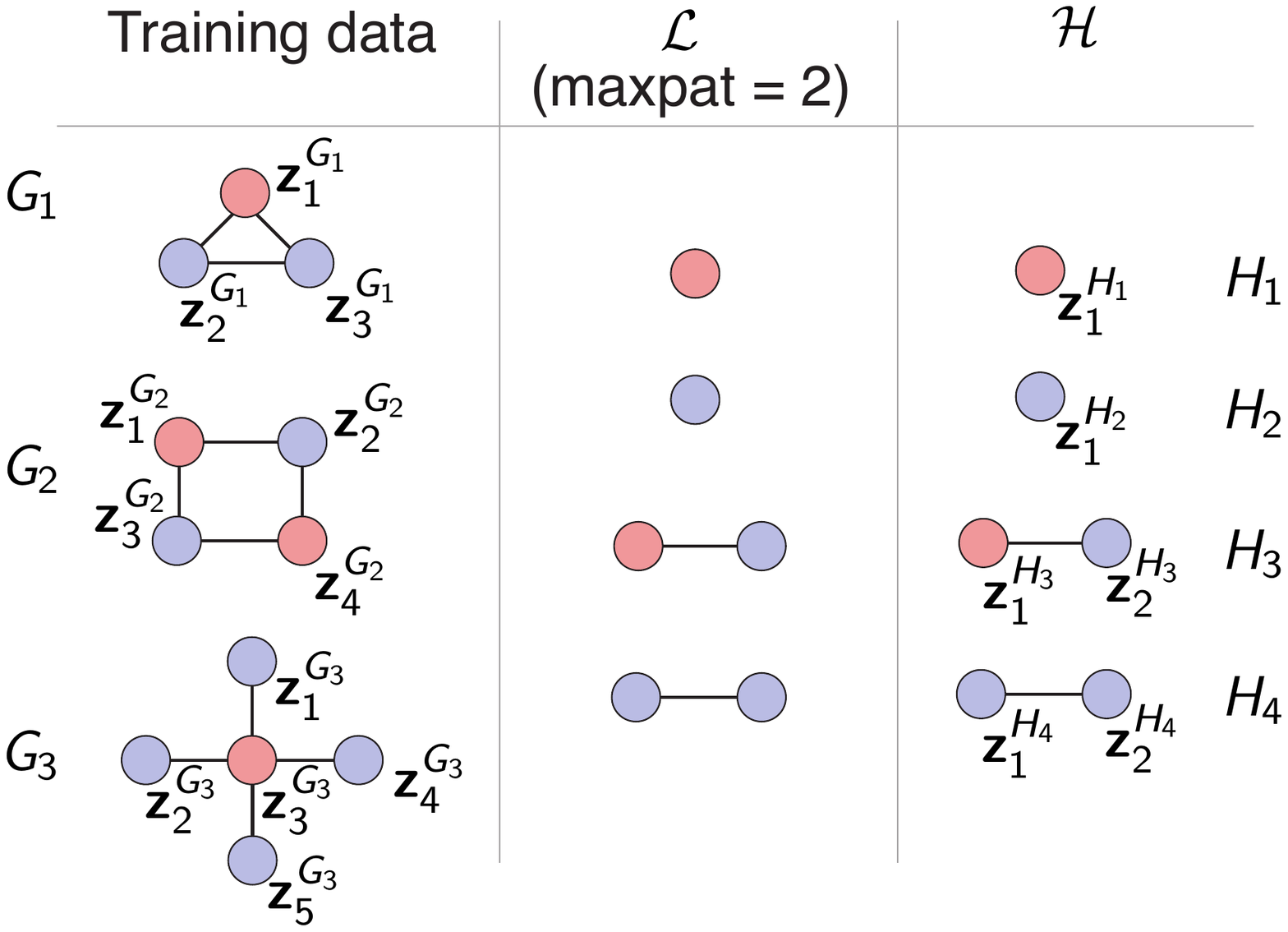}
 \end{center}
 \caption{
 An example of training data, $\cL$ and $\cH$.
 Since $\cL$ only includes subgraphs in the training data, ``\includegraphics[width=.7cm]{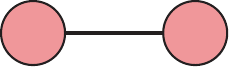}'' is not included in $\cL$.
 $\cH$ is created from $\cL$ by adding trainable attribute vectors $\*z^{H_i}_{v}$ ($v \in V_{H_i}$).
 }
 \label{fig:dataset}
\end{figure}

Our prediction model linearly combines the feature $\psi(G_i ; H)$ as follows: 
\begin{align}
 f(G_i) = \sum_{H \in \cH} \psi(G_i ; H) \beta_{H} + \beta_0 = \*\psi_i^\top \*\beta + \beta_0, 
 \label{eq:f}
\end{align}
where $\beta_{H}$ and $\beta_0$ are parameters, $\cH$ is a set of candidate AGs, and $\*\beta$ and $\*\psi_i$ are vectors containing $\beta_H$ and $\psi(G_i ; H)$ for $H \in \cH$, respectively.
Let 
$\cL = \{ L \mid L \subseteq L(G_i), i \in [n], |L| \leq \text{maxpat}\}$
be a set of all the labeled subgraphs contained in the training input graphs $\{ G_i \}_{i \in [n]}$, where $|L|$ is the number of nodes in the labeled graph $L$ and $\text{maxpat}$ is the user-specified maximum size of AGs.
The number of the candidate AGs $|\cH|$ is set as the same size as $|\cL|$.
%
We set $\cH$ as a set of attributed graphs created by giving trainable attribute vectors $\*z_v^H (v \in V_H)$ to each one of elements in $\cL$. 
\figurename~\ref{fig:dataset} shows a toy example.
Our optimization problem for 
$\beta_H$, $\beta_0$ and $\*z_v^H$ 
is defined as the following regularized loss minimization in which the sparse $L1$ penalty is imposed on $\beta_H$: 
\begin{align}
 \min_{\*\beta, \beta_0, \cZ_\cH}
 \
 \frac{1}{2}
 \sum_{i = 1}^n
 \ell(y_i, f(G_i))
 + \lambda \sum_{H \in \cH} | \beta_H |, 
 \label{eq:objective}
\end{align}
where 
$\cZ_\cH = \{ \*z^H_v \mid v \in V_H, H \in \cH \}$, 
and
$\ell$ is a convex differentiable loss function.
Here, we employ the squared hinge loss function \cite{janocha2016loss}:
\begin{align*}
 \ell(y_i, f(G_i)) = \max(1 - y_i f(G_i), 0)^2. 
\end{align*}
Since the objective function \eq{eq:objective} induces a sparse solution for $\beta_H$, we can identify a small number of important AGs as $H$ having the non-zero $\beta_H$.
However, this optimization problem has an intractably large number of optimization variables ($\cH$ contains all the possible subgraphs in the training dataset and each one of $H \in \cH$ has the attribute vector $\*z_v^H \in \RR^d$ for each one of nodes).
We propose an efficient optimization algorithm that mitigates this problem.

\subsection{Optimization}
\label{ssec:optimization}

Our optimization algorithm is based on the block coordinate update \cite{xu2017globally} algorithm, in which the (proximal) gradient descent alternately updates a block of variables.
%
We update one of $\*\beta, \beta_0$ and $\cZ_\cH$ alternately, while the other two parameters are fixed.
First, the proximal gradient update is applied to $\*\beta$ because it has the $L1$ penalty.
Second, for $\beta_0$, we calculate the optimal solution under fixing the other variable because it is easy to obtain.
Third, for $\cZ_\cH$, we apply the usual gradient descent update because it does not have sparse penalty.
%

The difficulty of the optimization problem \eq{eq:objective} originates from the size of $\cH$.
We select a small size of a subset 
$\cW \subseteq \cH$, 
and only  
$\*\beta_{\cW} \in \RR^{|\cW|}$, 
defined by $\beta_H$ for $H \in \cW$,
and corresponding attribute vectors 
$\cZ_\cW = \{ \*z^H_v \mid v \in V_H, H \in \cW \} \subseteq \cZ_\cH$
are updated.
We propose an efficient pruning strategy by combining the proximal gradient with the graph mining, which enables us to select $\cW$ without enumerating all the possible subgraphs.
A notable characteristics of this approach is that it can obtain the completely same result compared with when we do not restrict the size of variables.

\subsubsection{Update $\*\beta, \beta_0$ and $\cZ_\cH$} 
\label{sssec:update}

Before introducing the subset $\cW$, we first describe update rules of each variable. 
First, we apply the proximal gradient update to $\*\beta$.
Let 
\begin{align*}
 g_H(\*\beta) = 
 \sum_{i = 1}^n \pd{\ell(y_i, f(G_i))}{f(G_i)} \psi(G_i ; H)
\end{align*}
be the derivative of the loss term in \eq{eq:objective} with respect to $\beta_H$.
Then, the update of $\*\beta$ is defined by
\begin{align}
 \beta^{\rm (new)}_H \leftarrow 
 \mathrm{prox} 
 \left(
 \beta_H - \eta \ g_H(\*\beta)
 \right),
 \label{eq:update-beta}
\end{align}
where $\eta > 0$ is a step length, and 
\begin{align*}
 \mathrm{prox}(a) = 
 \begin{cases}
  a - \eta \lambda & \text{ if } a \geq \eta \lambda, \\
  0 & \text{ if } a \in (- \eta \lambda, \eta \lambda), \\
  a + \eta \lambda & \text{ if } a \leq - \eta \lambda
 \end{cases}
\end{align*}
is a proximal operator (Note that the proximal gradient for the $L1$ penalty is often called ISTA \cite{bech2009fast}, for which an accelerated variant called FISTA is also known. We here employ ISTA for simplicity).
We select the step length $\eta$ by the standard backtrack search.

The bias term $\beta_0$ is update by 
\begin{align*}
 \min_{\beta_0} \ \sum_{i = 0}^n \max(1 - y_i (\*\psi_i^\top \*\beta + \beta_0), 0)^2,
\end{align*}
which is the optimal solution of the original problem \eq{eq:objective} for given other variables $\*\beta$ and $\cZ_\cH$.
Since the objective function of $\beta_0$ is a differential convex function, the update rule of $\beta_0$ can be derived from the first order condition as 
\begin{align}
 \beta^{\rm (new)}_0 \leftarrow
 \frac{ \sum_{i \in \cI^{\rm (new)}}  (y_i - \*\psi_i^\top \*\beta) }{ | \cI^{\rm (new)} | },
 \label{eq:update-beta0}
\end{align}
where  
$\cI^{\rm (new)} = \{i \mid 1 - y_i(\*\psi_i^\top \*\beta + \beta^{\rm (new)}_0) > 0\}$.
This update rule contains $\beta^{\rm (new)}_0$ in $\cI^{\rm (new)}$. 
%
%
However, it is easy to calculate the update \eq{eq:update-beta0} without knowing $\beta^{\rm (new)}_0$ beforehand.  
%
Here, we omit detail because it is a simple one dimensional problem (see supplementary appendix~\ref{app:update-beta0}).

For $\*z^H_v \in \cZ_{\cH}$, we employ the standard gradient descent:
\begin{align}
 \*z^{H^{\rm (new)}}_v \leftarrow \*z^H_v - \alpha
 \sum_{i = 1}^n \pd{\ell(y_i, f(G_i))}{f(G_i)} 
 \times
 \left(
 \beta_{H} 
 \pd{\psi(G_i ; H)}{\*z^H_v}
 \right), 
 \label{eq:update-z}
\end{align}
where $\alpha > 0$ is a step length to which we apply the standard backtrack search.

\subsubsection{Gradient Pruning with Graph Mining}
\label{sssec:pruning}

In every update of $\*\beta$, we incrementally add required $H$ into $\cW \subseteq \cH$.
For the complement set 
$\overline{\cW} = \cH \setminus \cW$, 
which contains AGs that have never been updated, we initialize
$\beta_H = 0$ for $H \in \overline{\cW}$.
For the initialization of a node attribute vector
$\*z_v^H \in \cZ_\cH$, 
we set the same initial vector if the node (categorical) labels are same, i.e., 
$\*z_v^H = \*z_{v^\prime}^{H^\prime}$ if $L_v = L_{v^\prime}$ for $\forall H, H^{\prime} \in \cH$
(in practice, we use the average of the attribute vectors within each node label).
This constraint is required for our pruning criterion, but it is only for initial values. After the update \eq{eq:update-z}, all $\*z^H_v$ can have different values.

Since $\beta_H = 0$ for $H \in \overline{\cW}$, it is easy to derive the following relation from the proximal update \eq{eq:update-beta}:
\begin{align}
 |g_H(\*\beta)| \leq \lambda \text{ and } H \in \overline{\cW}
 \ \Rightarrow \
 0 = \mathrm{prox} \left(
 \beta_H - \eta \ g_H(\*\beta)
 \right).	
 \label{eq:gradient-condition}
\end{align}
This indicates that if the conditions in the left side hold, we do not need to update $\beta_H$ because it remains $0$.
Therefore, we set 
\begin{align}
 \cW \leftarrow \cW \cup \left\{ H \ \middle| \ |g_H(\*\beta)| > \lambda, \forall H \in \overline{\cW} \right\}, 
 \label{eq:update-W}
\end{align}
and apply the update \eq{eq:update-beta} only to $H \in \cW$.
However, evaluating 
$|g_H(\*\beta)| > \lambda$
for all 
$H \in \overline{\cW}$
can be computationally intractable because it needs to enumerate all the possible subgraphs.
The following theorem can be used to avoid this difficulty:
\begin{theo} \label{th:pruning}
 Let $L(H^\prime) \sqsupseteq L(H)$ and $H, H^\prime \in \overline{\cW}$.
 Then, 
 \begin{align*}
  |g_{H^\prime}(\*\beta)| \leq \overline{g}_H(\*\beta)
 \end{align*}
 where
 \begin{align*}
  \overline{g}_H(\*\beta)
  = 
  \max \Biggl\{ 
  & \sum_{ i \in {\cI \cap} \{ i \mid y_i > 0 \}}
  y_i \psi(G_i; H) (1 - y_i (\*\psi_i^\top \*\beta + \beta_0)), \\
  & - \sum_{ i \in {\cI \cap} \{ i \mid y_i < 0 \}}
  y_i \psi(G_i; H) (1 - y_i (\*\psi_i^\top \*\beta + \beta_0))
  \Biggr\},
 \end{align*}
 where $\cI = \{i \mid 1 - y_i(\*\psi_i^\top \*\beta + \beta_0) > 0\}$.
\end{theo}
\noindent
See supplementary appendix~\ref{app:proof-theorem2-1} for the proof.
Note that here $\cI$ is defined by the current $\beta_0$ unlike \eq{eq:update-beta0}.
This theorem indicates that the gradient 
$|g_{H^\prime}(\*\beta)|$
for any $H^\prime$ whose $L(H^\prime)$ contains $L(H)$ as a subgraph can be bounded by 
$\overline{g}_H(\*\beta)$.
%
It should be noted that 
$\overline{g}_H(\*\beta)$
can be calculated without generating $H^\prime$, 
\red{ and it mainly needs only the model prediction with the current parameter
$\*\psi_i^\top \*\beta + \beta_0$, which can be immediately obtained at each iteration,  
and AGIS $\psi(G_i; H)$.}
The rule \eq{eq:gradient-condition} reveals that, to identify $\beta_{H^\prime}^{\rm (new)} = 0$, we only require to know whether $|g_{H^\prime}(\*\beta)| \leq \lambda$ holds, and thus, an important consequence of theorem~\ref{th:pruning} is the following rule:
\begin{align}
 \begin{split}  
 & \overline{g}_H(\*\beta) 
  \leq \lambda \text{ and } H \in \overline{\cW}	 \\
 & \ \Rightarrow \
  |g_{H^\prime}(\*\beta)| \leq \lambda 
  \text{ for } 
  \forall H^\prime \in \{ H^\prime \mid L(H^\prime) \sqsupseteq L(H), H^\prime \in \overline{\cW} \}.	
 \end{split}
 \label{eq:pruning}
\end{align}
Therefore, if the conditions in the first line in \eq{eq:pruning} hold, any $H^\prime$ whose $L(H^\prime)$ contains $L(H)$ as a subgraph can be discarded during that iteration.
Further, from \eq{eq:update-z}, we can immediately see that attribute vectors
$\*z^H_v$ for $\forall v \in V_H$
are also not necessary to be updated if $\beta_H = 0$.
This is an important fact because updates of a large number of variables can be omitted.

\begin{figure*}
 \begin{center}
  \igr{.65}{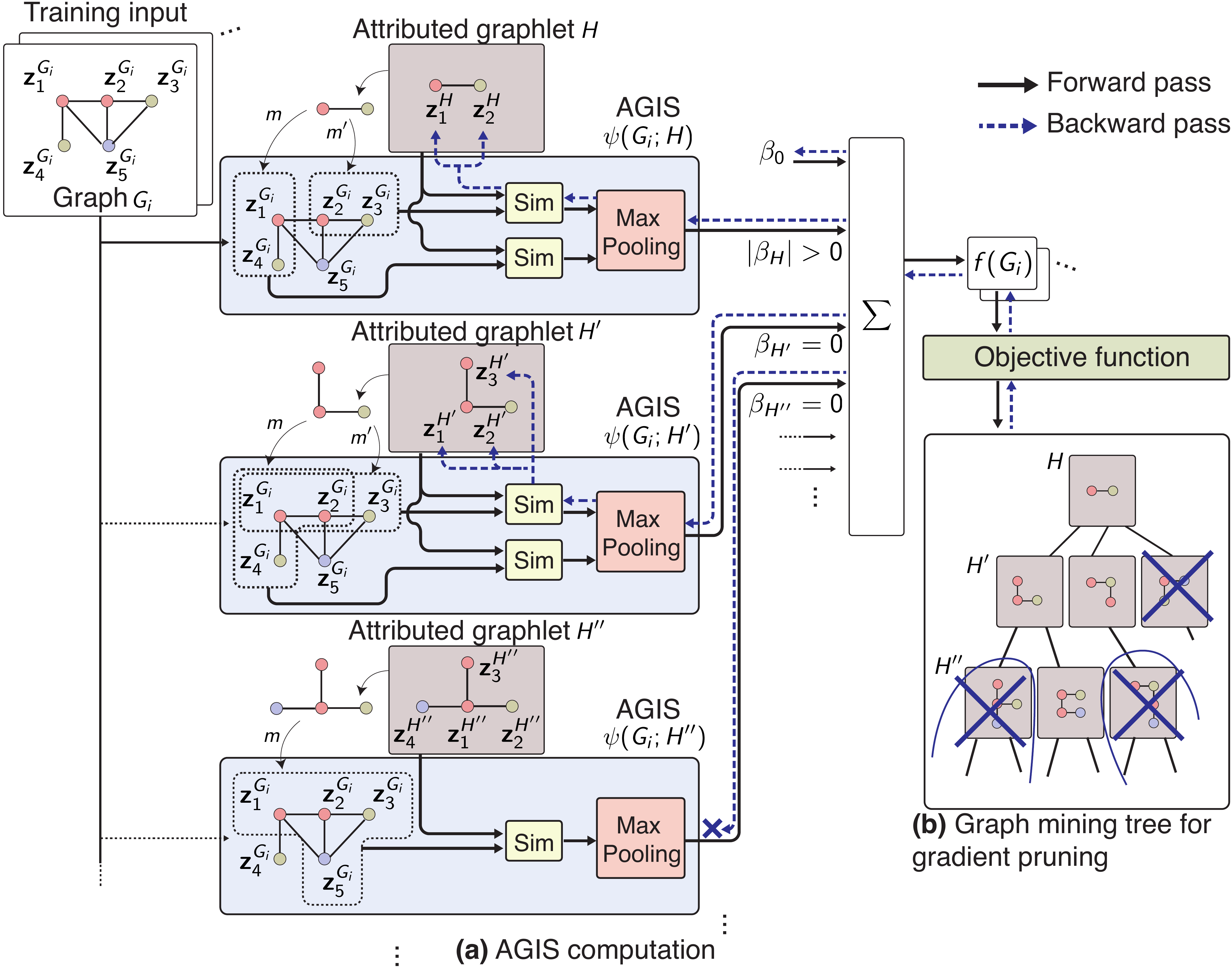}
 \end{center}
 \caption{\red{
 An illustration of LAGRA.
 a)
 In the forward pass, only passes with $|\beta_H| > 0$ contribute to the output. 
 AGIS is defined by the best matching between an input graph and an AG.
 b)
 For the backward pass, the gradient can be pruned when the rule \eqref{eq:pruning} is satisfied.
 In this illustration, $H^{\prime\prime}$ is pruned by which graphs expanded from $H^{\prime\prime}$ are not required to compute the gradient.
 }}
 \label{fig:overview}
\end{figure*}

\red{\figurename~\ref{fig:overview} shows an illustration of the forward and backward (gradient) computations of LAGRA.
For the gradient pruning, an efficient algorithm can be constructed by combining the rule \eq{eq:pruning} and a graph mining algorithm.
}
%
%
%
A well-known efficient graph mining algorithm is gSpan \cite{yan2002gspan}, which creates the tree by recursively expanding each graph in the tree node as far as the expanded graph is included in a given set of graphs as a subgraph. 
%
An important characteristics of \red{the mining tree} is that all graphs must contain any graph of its ancestors as subgraphs.
Therefore, during the tree traverse (depth-first search) by gSpan, we can prune the entire subtree (all descendant nodes) if 
$\overline{g}_H(\*\beta) \leq \lambda$
holds for the AG $H$ in a tree node \red{(\figurename~\ref{fig:overview}(b))}.
This means that we can update $\cW$ by \eq{eq:update-W} without exhaustively investigating all the elements in $\overline{\cW}$.


gSpan has another advantage for LAGRA.
To calculate the feature $\phi_H(G_i)$, defined in \eq{eq:phi}, LAGRA requires a set of injections $M$ (an example is shown in \figurename~\ref{fig:graph-matching}(b)). 
gSpan keeps the information of $M$ during the tree traverse because it is required to expand a subgraph in each $G_i$ (see the authors implementation \url{https://sites.cs.ucsb.edu/~xyan/software/gSpan.htm}).
%
Therefore, we can directly use $M$ created by gSpan to calculate \eq{eq:phi}.

\subsubsection{Algorithm}
\label{ssec:algorithm}

We here describe entire procedure of the optimization of LAGRA.
We employ the so-called regularization path following algorithm (e.g., \cite{friedman2007pathwise}), in which the algorithm starts from a large value of the regularization parameter $\lambda$ and gradually decreases it while solving the problem for each $\lambda$.
This strategy can start from highly sparse $\*\beta$, in which usually $\cW$ also becomes small. 
Further, at each $\lambda$, the solution obtained in the previous $\lambda$, can be used as the initial value by which faster convergence can be expected (so-called warm start).

Algorithm~\ref{alg:regularization-path} shows the procedure of the regularization path following.
\red{
We initialize $\*\beta = \*0$, which is obviously optimal when $\lambda = \infty$.
In line 2 of Algorithm~\ref{alg:regularization-path}, we calculate $\lambda_{\max}$ at which $\*\beta$ starts having non-zero values:
%
$\lambda_{\max} = \max_{H \in \cH} 
 \left| 
 \sum_{i \in [n]}^n y_i \psi(G_i ; H) (1 - y_i \beta_0)
 \right|$,
where $\beta_0 = \sum_{i \in [n]} y_i / n$. 
%
See supplementary appendix~\ref{app:lambda_max} for derivation.
%
$\lambda_{\rm max}$ can also be written as $\lambda_{\max} = \max_{H \in \cH} |g_H(\*0)|$. 
To find $\max_{H \in \cH}$, we can use almost the same gSpan based pruning strategy by using an upper bound of 
$\overline{g}_H(\*\beta)$
as shown in Section~\ref{sssec:pruning} (the only difference is to search the $\max$ value only, instead of searching all $H$ satisfying $|g_H(\*\beta)| > \lambda$), though in Algorithm~\ref{alg:regularization-path}, this process is omitted for brevity.}
%
After setting $\lambda_0 \leftarrow \lambda_{\max}$, the regularization parameter $\lambda$ is decreased by using a pre-defined decreasing factor $R$ as shown in line 6 of Algorithm~\ref{alg:regularization-path}.
For each $\lambda_1 > \cdots > \lambda_K$, the parameters $\*\beta, \beta_0$ and $\cZ_\cH$ are alternately updated as described in Section~\ref{sssec:update} and \ref{sssec:pruning}. 
We stop the alternate update by monitoring performance on the validation dataset in line 14 (stop by thresholding the decrease of the objective function is also possible).

The algorithm of the pruning strategy described in Section~\ref{sssec:pruning} is shown in Algorithm~\ref{alg:gpruning}.
This function recursively traverses the graph mining tree.
At each tree node, first, 
$\overline{g}_H(\*\beta)$
is evaluated to prune the subtree if possible. 
Then, if $|g_H(\*\beta)| > \lambda_k$, $H$ is included in $\cW$. 
The expansion from $H$ (creating children of the graph tree) is performed by gSpan, by which only the subgraphs contained in the training set can be generated (\red{see the original paper \cite{yan2002gspan} for detail of gSpan}).
The initialization of the trainable attribute 
$\*z_v^{H^\prime}$ 
is performed when 
$H^\prime$ 
is expanded (line 15).

\begin{algorithm}
    \caption{Optimization of LAGRA} \label{alg:regularization-path}
    \Function(Reguralization-Path{(}$K,R,\mathrm{MaxEpoch}${)}){
    $H_0 \leftarrow \text{a graph at the root node of the mining tree}$ \\
    $\mathcal{W} \leftarrow \emptyset$\\ 
    $\*\beta \leftarrow \*0, \ \beta_0 = \sum_{i \in [n]} y_i / n$\\
    $\lambda_0\leftarrow\lambda_{\max}$ \Comment{Compute $\lambda_{\max}$}
    \For{$k=1,2,\ldots, K$}{
    $\lambda_k \leftarrow R\lambda_{k-1}$\\
    \For{$\mathrm{epoch}=1,2,\ldots, \mathrm{MaxEpoch}$}{
    $\mathcal{W} \leftarrow \mathcal{W} \cup \mathrm{GradientPruning}(H_0, ~\lambda_k)$\\
    Update $\*\beta$ by \eq{eq:update-beta} for $H \in \cW$\\
    Update $\beta_0$ by \eq{eq:update-beta0} \\
    Update $\*z^H_v$ by \eq{eq:update-z} for $H \in \cW$\\
    $\mathrm{val\_loss}\leftarrow$ Compute validation loss\\
    \If{ {\rm val\_loss has not been improved in the past $q$ iterations} }{
    break \Comment{Inner loop stopping condition}
    }
    \Else{
        $\mathcal{M}^{(k)} \leftarrow (\mathcal{W}, \bm\beta, \beta_0)$\\
    }
    }
    }
    \Return{$\{\mathcal{M}^{(k)}\}_{k=0}^{K}$}
    }
\end{algorithm}
%
\begin{algorithm}
 \caption{Gradient Pruning} \label{alg:gpruning}
    \Function(GradientPruning{(}$H, ~\lambda_k${)}){
            $\mathcal{W} \leftarrow \emptyset$\\
            \If{$\overline{g}_H(\*\beta) \leq \lambda_k$}{ 
                \Return{$\emptyset$} \Comment{Prune the subtree}
            }
            \If{$| g_H(\*\beta) | > \lambda_k$}{
                $\mathcal{W} \leftarrow \mathcal{W} \cup \{ H \}$
            }
            $\cC \leftarrow \mathrm{CreateChildren}(H)$\\
            \For{$H^\prime \in \cC$}{
                $\mathcal{W} \leftarrow \mathcal{W} \cup \mathrm{GradientPruning}(H^\prime, ~\lambda_k)$
            }
            \Return{$\mathcal{W}$}
        }
    \Function(CreateChildren{(}$H${)}){
            \If{ {\rm children of $H$ have never been created by gSpan} }{
                $\cC \leftarrow$ graphs expanded from $H$ by gSpan\\
                \For{ $H^\prime \in \cC$ }{
                    $\*z_v^{H^\prime} \leftarrow {\rm mean} \{\*z_{v^\prime}^{G_i} \mid v^\prime \in V_{G_i}, L_{v} = L_{v^\prime}, i \in [n] \}$\\
                    Compute $\{ \psi(G_i ; H^\prime) \}_{i=1}^n$ using $M$ created by gSpan
                }
            }
        }
\end{algorithm}

\begin{table*}
 \centering
 \caption{
 Classification accuracy and the number of selected non-zero $\beta_H$ by LAGRA (the bottom row).
 The average of five runs and its standard deviation are shown.
 The underlines indicate the best average accuracy for each dataset and the bold-face indicates that the result is comparable with the best method in a sense of one-sided $t$-test (significance level 5\%).
 \# best indicates frequency that the method is the best or comparable with the best method.
 } \label{tab:accuracy}
	\scalebox{0.6}{
		\begin{tabular}{|c|c|c|c|c|c|c|c||c|}
			\hline
 			          & AIDS                       & BZR                        & COX2                       & DHFR                       & ENZYMES                  & PROTEINS                   & SYNTHETIC  & \# best            \\ \hline \hline
GH & \underline{\textbf{0.9985 $\pm$ 0.0020}} & 0.8458 $\pm$ 0.0327 & 0.7872 $\pm$ 0.0252 & 0.7250 $\pm$ 0.0113 & 0.6050 $\pm$ 0.0857 & \textbf{0.7277 $\pm$ 0.0332} & 0.6767 $\pm$ 0.0655 & 2  \\  \hline 
ML & 0.9630 $\pm$ 0.0062 & 0.8289 $\pm$ 0.0141 & 0.7787 $\pm$ 0.0080 & 0.7105 $\pm$ 0.0300 & 0.6000 $\pm$ 0.0652 & 0.6205 $\pm$ 0.0335 & 0.4867 $\pm$ 0.0356 & 0\\  \hline 
PA & 0.9805 $\pm$ 0.0086 & 0.8313 $\pm$ 0.0076 & 0.7809 $\pm$ 0.0144 & 0.7316 $\pm$ 0.0435 & \underline{\textbf{0.7500 $\pm$ 0.0758}} & 0.6884 $\pm$ 0.0077 & 0.5400 $\pm$ 0.0859 & 1 \\  \hline 
DGCNN & 0.9830 $\pm$ 0.0046 & 0.8169 $\pm$ 0.0177 & \textbf{0.8021 $\pm$ 0.0401} & 0.7289 $\pm$ 0.0192 & \textbf{0.7289 $\pm$ 0.0192} & \underline{\textbf{0.7509 $\pm$ 0.0114}} & 0.9867 $\pm$ 0.0125 & 3\\  \hline 
GCN & 0.9840 $\pm$ 0.0030 & 0.8290 $\pm$ 0.0460 & \underline{\textbf{0.8340 $\pm$ 0.0257}} & 0.7490 $\pm$ 0.0312 & \textbf{0.7000 $\pm$ 0.0837} & 0.6880 $\pm$ 0.0202 & 0.9630 $\pm$ 0.0194 & 2 \\  \hline 
GAT & 0.9880 $\pm$ 0.0041 & 0.8220 $\pm$ 0.0336 & 0.7830 $\pm$ 0.0274 & 0.7110 $\pm$ 0.0156 & \textbf{0.7100 $\pm$ 0.0768} & 0.7160 $\pm$ 0.0108 & \textbf{0.9800 $\pm$ 0.0267} & 2 \\  \hline  
LAGRA (Proposed) & 0.9900 $\pm$ 0.0050 & \underline{\textbf{0.8892 $\pm$ 0.0207}} & 0.8043 $\pm$ 0.0229 & \underline{\textbf{0.8171 $\pm$ 0.0113}} & 0.6450 $\pm$ 0.0797 & \textbf{0.7491 $\pm$ 0.0142} & \underline{\textbf{1.0000 $\pm$ 0.0000}} & 4 \\  \hline  \hline 
\# non-zero $\beta_H$&   50.4 $\pm$ 17.1 & 52.4 $\pm$ 19.0 & 45.4 $\pm$ 14.9 & 40.0 $\pm$ 11.6 & 7.2 $\pm$ 8.4 & 25.8 $\pm$ 9.4 & 35.8 $\pm$ 35.0 & - \\ \hline
		\end{tabular}
	}
\end{table*}

\section{Related Work}
\label{sec:related-work}

For graph-based prediction problems, recently, graph neural networks (GNNs) \cite{zhou2020graph} have attracted wide attention. 
However, interpreting GNNs is not easy in general.
According to a recent review of explainable GNNs \cite{yuan2022explainability}, almost all of explainability studies for GNNs are instance-level explanations, which provides input-dependent explanations (Here, we do not mention each one of input-dependent approaches because the purpose is clearly different from LAGRA).
An exception is XGNN \cite{yuan2020xgnn}, in which important discriminative graphs are generated for a given already trained GNN by maximizing the GNN output for a target label. 
%
However, unlike our method, the prediction model itself remains black-box, and thus, \red{it is difficult to know underlying dependency between the identified graphs and the prediction}. 

A classical approach to graph-based prediction problems is the graph kernel \cite{kriege2020survey}. 
%
%
%
Although graph kernel itself does not identify important substructures, recently, \cite{Feng2022kergnns} has proposed an interpretable kernel-based GNN, called KerGNN.
KerGNN uses a graph kernel function as a trainable filter, inspired by the well-known convolutional networks, and the filter updates the node attributes of the input graph so that it embeds similarity to learned important subgraphs.
Then, \cite{Feng2022kergnns} claims that resulting graph filter can be seen as a key structure.
%
\red{However, a learned subgraph in a graph kernel filter is difficult to interpret.
The kernel-based matching does not guarantee the existence of a subgraph unlike our AGIS \eq{eq:AGIS}, and further, only 1-hop neighbors of each node in the input graph are matched to a graph filter.
}
%
%
%




Another graph mining based approach is \cite{nakagawa2016safe}.
%
This approach also uses a pruning based acceleration for the optimization, but it is based on the optimality of the convex problem while our proximal gradient pruning is applicable to the non-convex problem of LAGRA. 
Further, more importantly, \cite{nakagawa2016safe} cannot deal with continuous attributes. 
%
The prediction model of LAGRA is inspired by a method for learning time-series shaplets (LTS) \cite{grabocka2014learning}. 
LTS is also based on a linear combination of trainable shaplets, which is a short fragment of a time-series sequence.
Unlike time-series data, possible substructures in graph data have a combinatorial nature because of which our problem setting has a computational difficulty that does not exist in the case of LTS, for which LAGRA provide a graph mining based efficient strategy.

\section{Experiments}
\label{sec:experiments}

Here, we empirically verify effectiveness of LAGRA.
We used standard graph benchmark datasets, called AIDS, BZR, COX2, DHFR, ENZYMES, PROTEINS and SYNTHETIC, retrieved from 
\url{https://ls11-www.cs.tu-dortmund.de/staff/morris/graphkerneldatasets}
(for ENZYMES, we only used two classes among original six classes to make a binary problem).
To simplify comparison, we only used node labels and attributes, and did not use edge labels and attributes.
Statistics of datasets are summarized in supplementary appendix~\ref{app:statistics-datasets}. 
The datasets are randomly divided into 
$\mathrm{train}:\mathrm{validation}:\mathrm{test}=0.6:0.2:0.2$.
%
%
For the regularization path algorithm (Algorithm~\ref{alg:regularization-path}), we created candidate values of $\lambda$ by uniformly dividing $[\log(\lambda_{\max}), \log(0.01\lambda_{\max})]$ into $100$ grid points.
We selected $\lambda$, $\text{maxpat} \in \{5,10\}$ and $\rho \in \{ 1, 0.5, 0.1, 0.05, 0.01 \}$ based on the validation performance.

\subsection{Prediction Performance}
\label{ssec:prediction-performance}

For the prediction accuracy comparison, we used graph kernels and graph neural networks (GNN). 
We used three well-known graph kernels that can handle continuous attributes, i.e., graph hopper kernel (GH) \cite{feragen2013scalable}, multiscale Laplacian kernel (ML) \cite{kondor2016multiscale} and propagation kernel (PA) \cite{neumann2016propagation}, for all of which the library called GraKeL \cite{siglidis2020grakel} was used.
For the classifier, we employed the $k$-nearest neighbor ($k$-NN) classification for which each kernel function $k(G_i,G_j)$ defines the distance function as
$\| G_i - G_j \| = \sqrt{ k(G_i,G_i) - 2 K(G_i,G_j) + k(G_j,G_j) }$. 
The number of neighbors $k$ is optimized by the validation set.
For GNN, we used deep graph convolutional neural network (DGCNN) \cite{zhang2018end}, graph convolutional network (GCN) \cite{kipf2017semi}, and graph attention network (GAT) \cite{velickovic2018graph}. 
For DGCNN, the number of hidden units $\{ 64, 128, 256 \}$ and epochs are optimized by the validation set.
The other settings were in the default settings of the authors implementation 
\url{https://github.com/muhanzhang/pytorch_DGCNN}.
For GCN and GAT, we also selected the number of hidden units and epochs as above.
For other settings, we followed \cite{you2021identity}.

The results are shown in \tablename~\ref{tab:accuracy}. 
LAGRA was the best or comparable with the best method (in a sense of one-sided $t$-test) for BZR, DHFR, PROTEINS and SYNTHETIC (4 out of 7 datasets).
For AIDS and COX2, LAGRA has similar accuracy values to the best methods though they were not regarded as the best accuracy in $t$-test.
The three GNNs also show stable performance overall. 
Although our main focus is to build an interepretable model, we see that LAGRA achieved comparable accuracy with the current standard methods.
Further, LAGRA only used a small number of AGs shown in the bottom row of \tablename~\ref{tab:accuracy}, which suggests high interpretability of the learned models.


\begin{figure*}


%
\begin{center}  
 \igr{.98}{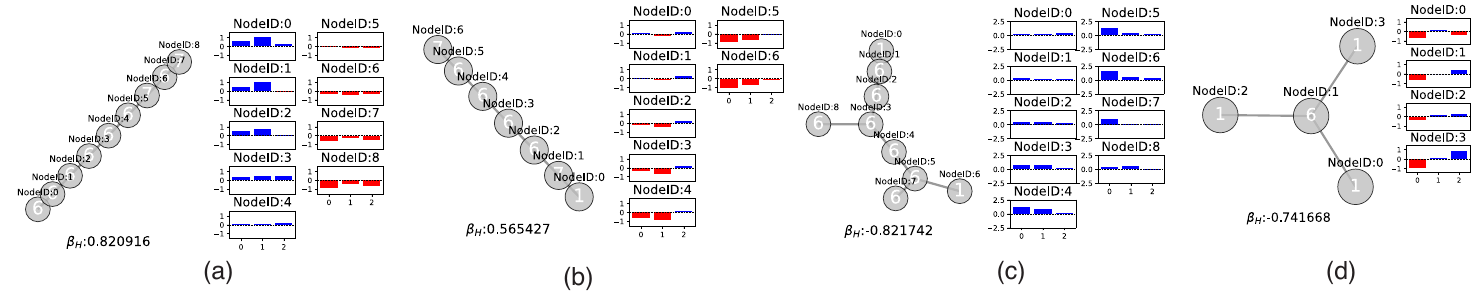}
\end{center}
 \caption{
 Selected important AGs for DHFR dataset.
 (a) and (b): AGs for the two largest positive coefficients. 
 (c) and (d): AGs for the two largest negative coefficients. 
 } \label{fig:AGs-DHFR}



\end{figure*}

\begin{figure*}
\begin{center}  
 \igr{.98}{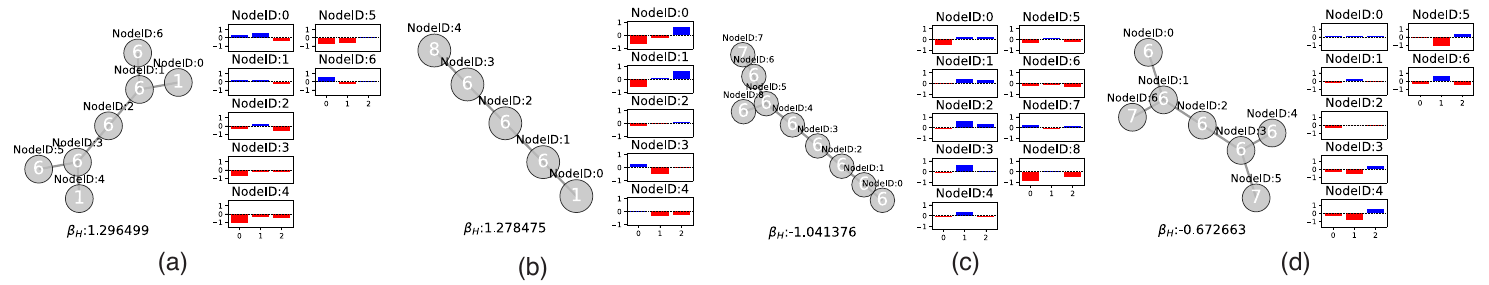}

\end{center}
 \caption{
 Selected important AGs for BZR dataset.
 (a) and (b): AGs for the two largest positive coefficients. 
 (c) and (d): AGs for the two largest negative coefficients. 
 } \label{fig:AGs-BZR}
\end{figure*}



\begin{figure}
 \begin{center}  
  \igr{.35}{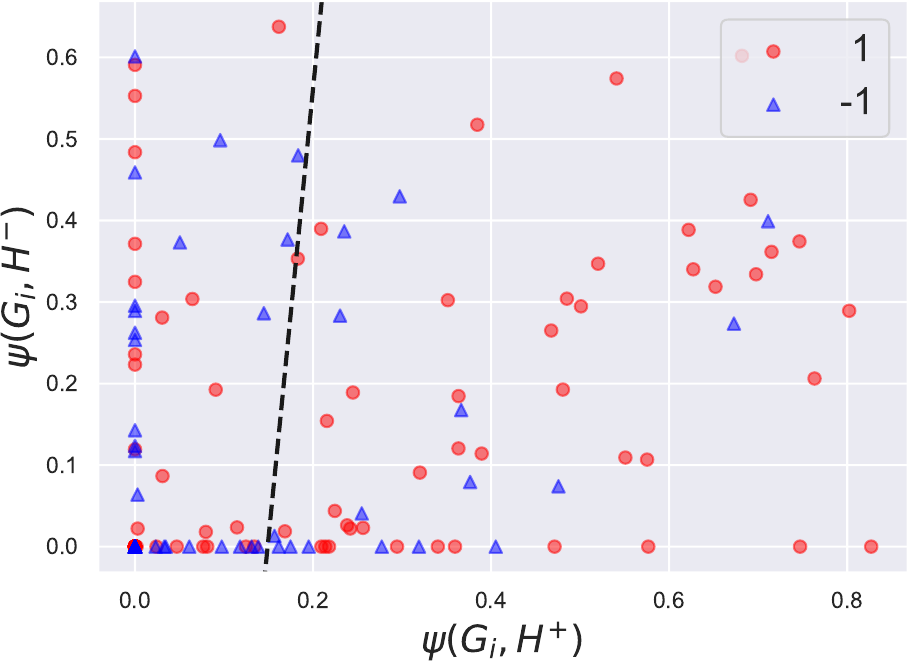}
 \end{center}
 \caption{
 \red{
 Scatter plots defined by selected AGs with test dataset of DHFR.
 The horizontal and vertical axes are AGIS of Fig.~\ref{fig:AGs-DHFR}(a) and Fig.~\ref{fig:AGs-DHFR}(c), respectively.}
 %
 }
 \label{fig:scatter}

\end{figure}

\subsection{Examples of Selected Attributed Graphlets}
\label{ssec:examples-selected-AG}

We here show examples of identified important AGs.
%
\red{\figurename~ \ref{fig:AGs-DHFR} and \ref{fig:AGs-BZR} show AGs having the two largest positive and negative $\beta_H$ for DHFR and BZR datasets, respectively.}
In each figure, a labeled graphlet $L(H)$ is shown in the left side (the numbers inside the graph nodes are the graph node labels) and {\it optimized} attribute vectors for each one of nodes are shown as bar plots in the right side.
We can clearly see important substractures not only by as structural information of a graph but also attribute values associated with each node.
%

Surprisingly, in a few datasets, two classes can be separated even in two dimensional space of AGIS.
\figurename~\ref{fig:scatter-AIDS} and \figurename~\ref{fig:scatter} show scatter plots of the test dataset (not the training dataset) with the axes of identified features by the LAGRA training. 
%
Let $H^+$ and $H^-$ be AGs having the largest positive and negative $\beta_H$, respectively.
The horizontal and vertical axes of plots are $\psi(G_i, H^+)$ and $\psi(G_i, H^-)$.
%
In particular, in the AIDS dataset, for which classification accuracy was very high in \tablename~\ref{tab:accuracy}, two classes are clearly separated.
For DHFR, we can also see points in two classes tend to be located on the upper left side and the lower right side.
%
\red{The dashed lines are boundaries created by (class-balance weighted) logistic regression fitted to the test points in these two dimensional spaces.}
The estimated class conditional probability has ${\rm AUC} = 0.94$ and $0.62$ for AIDS and BZR, respectively, which indicate that differences of two classes are captured even only by two AGs in these datasets.
%
%

\subsection{Discussion on Computational Time}
\label{ssec:computational-time}

Finally, we verify computational time of LAGRA. 
First, \tablename~\ref{tab:ws_max_size} shows the size of candidate AGs $|\cH|$ in each dataset.
As we describe in Section~\ref{sssec:model-definition}, this size is equal to $|\cL|$, i.e., the number of all the possible subgraphs in the training datasets.
Therefore, it can be quite large as particularly shown in the ENZYMES, PROTEINS and SYNTHETIC datasets in \tablename~\ref{tab:ws_max_size}.
The optimization variables in the objective function \eq{eq:objective} are $\*\beta, \beta_0$ and $\cZ_\cH$. 
The dimension of $\*\beta$ is $|\cH|$ and the node attribute vector $\*z_v^H \in \RR^d$ exists for each one of nodes in $H \in \cH$.
Thus, the number of optimization variables in \eq{eq:objective} is 
$1 + |\cH| + \sum_{H \in \cH} |H| \times d$, 
which can be prohibitively large.

\begin{table}
 \centering
 \caption{
 The size of the candidate set $\cH$ with maxpat 10.
 For the ENZYMES, PROTEINS, and SYNTHETIC datasets, only the lower bounds are shown because it took long time to count.
 }
 \label{tab:ws_max_size}
 \scalebox{0.8}{
 \begin{tabular}{|c|c|c|c|c|c|c|}
  \hline
  		 AIDS   & BZR    & COX2   & DHFR   & ENZYMES      & PROTEINS     & SYNTHETIC  
\\ \hline
  		 134281 & 148903 & 101185 & 137872 & $>$ 15464000 & $>$ 13987000 & $>$ 699000 
\\ \hline
 \end{tabular}
	}
\end{table}

\begin{figure}
 \begin{center}
  \subfloat[AIDS]{\igr{.3}{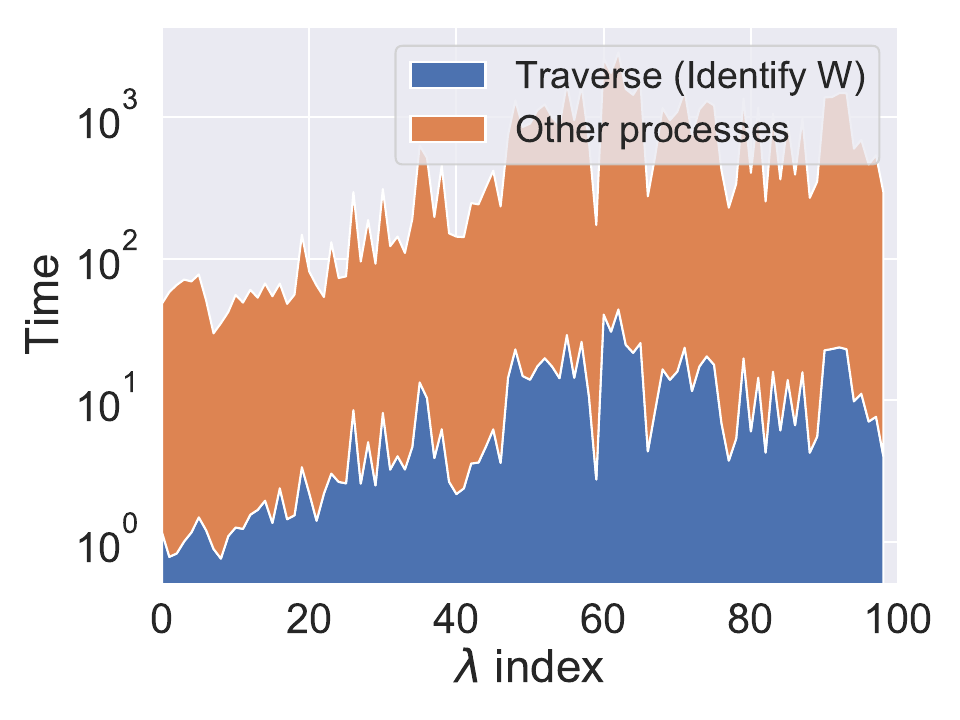}}
  %
  \subfloat[ENZYMES]{\igr{.3}{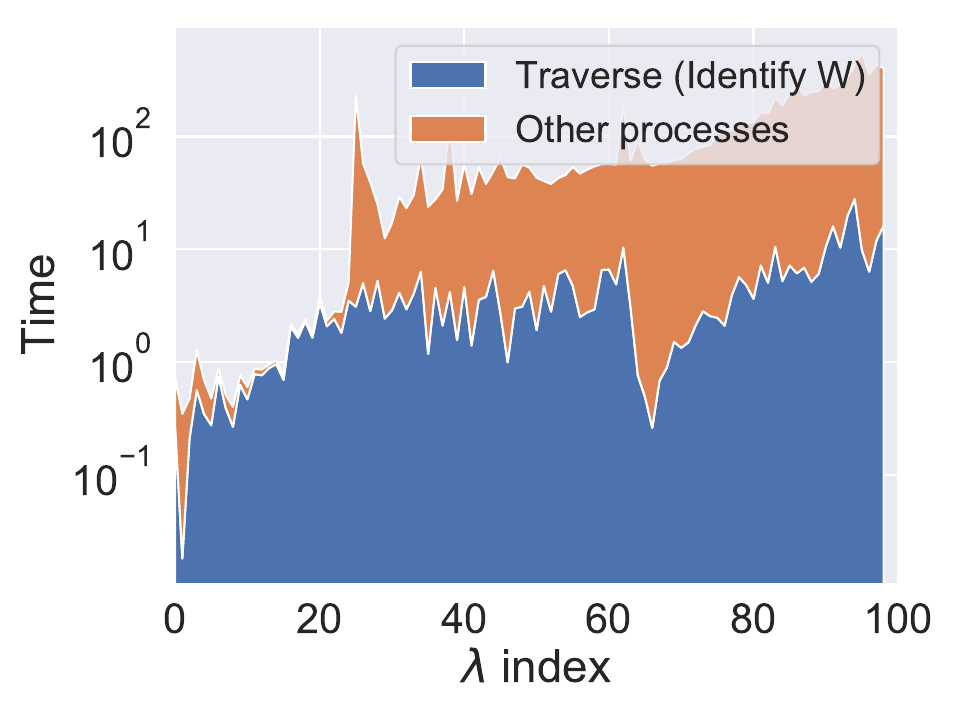}}
 \end{center}
 \caption{
 Transition of computational time (sec) on regularization path.
 For each $\lambda$, the total time and the time required to traverse the graph mining tree (in other words, the time required to identify $\cW$) is shown separately.
 }
 \label{fig:time-path}
\end{figure}

\figurename~\ref{fig:time-path} shows the computational time during the regularization path. 
The horizontal axis is $k$ of $\lambda_k$ in Algorithm~\ref{alg:regularization-path}. 
%
The datasets are AIDS and ENZYMES. 
In the regularization path algorithm, the number of non-zero $\beta_H$ typically increases during the process of decreasing $\lambda$, because the $L1$ penalty becomes weaker gradually.
As a results, in both the plots, the total time increases with the $\lambda$ index.
Although LAGRA performs the traverse of the graph mining tree in every iteration of the gradient update (line 9 in Algorithm~\ref{alg:regularization-path}), \figurename~\ref{fig:time-path} shows that the traverse time was not necessarily dominant (Note that the vertical axis is in log scale).
\red{
%
In particular, when only a small number of tree nodes are newly expanded at that $\lambda$, the calculation for the tree search becomes faster because AGIS $\psi(G_i,H)$ is already computed at the most of tree nodes.
}
%
The computational times were at most about $10^3$ sec for these datasets.
We do not claim that LGARA is computationally faster compared with other standard algorithms (such as graph kernels), but as the computational time of the optimization problem with
$1 + |\cH| + \sum_{H \in \cH} |H| \times d$
variables, the results obviously indicate effectiveness of our pruning based optimization approach.

\begin{figure}
 \begin{center}
  \subfloat[AIDS]{\igr{.3}{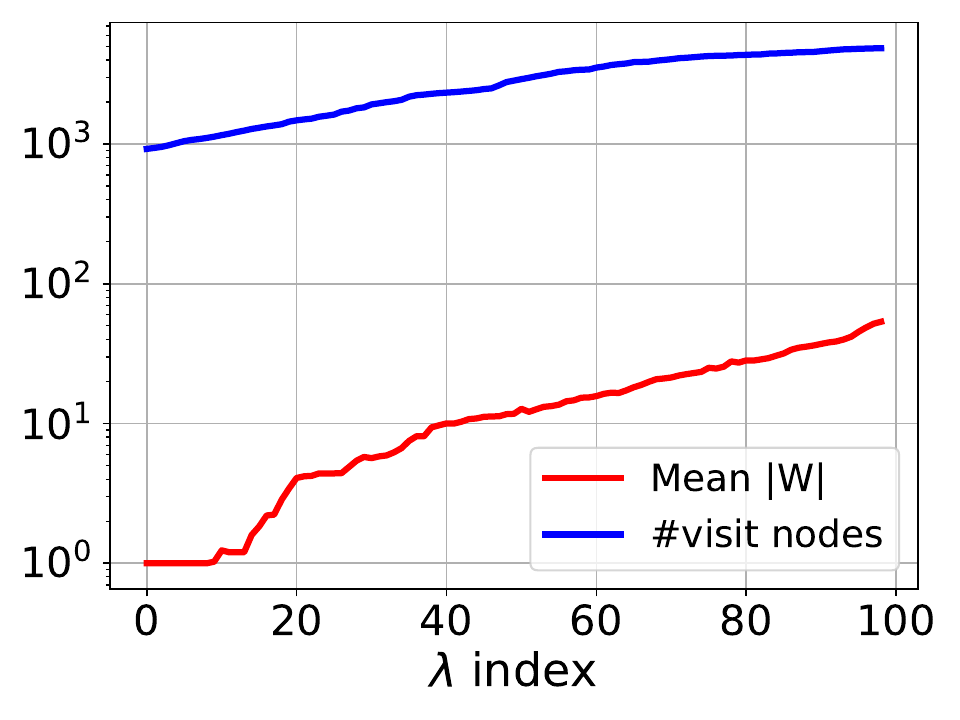}}
  \subfloat[ENZYMES]{\igr{.3}{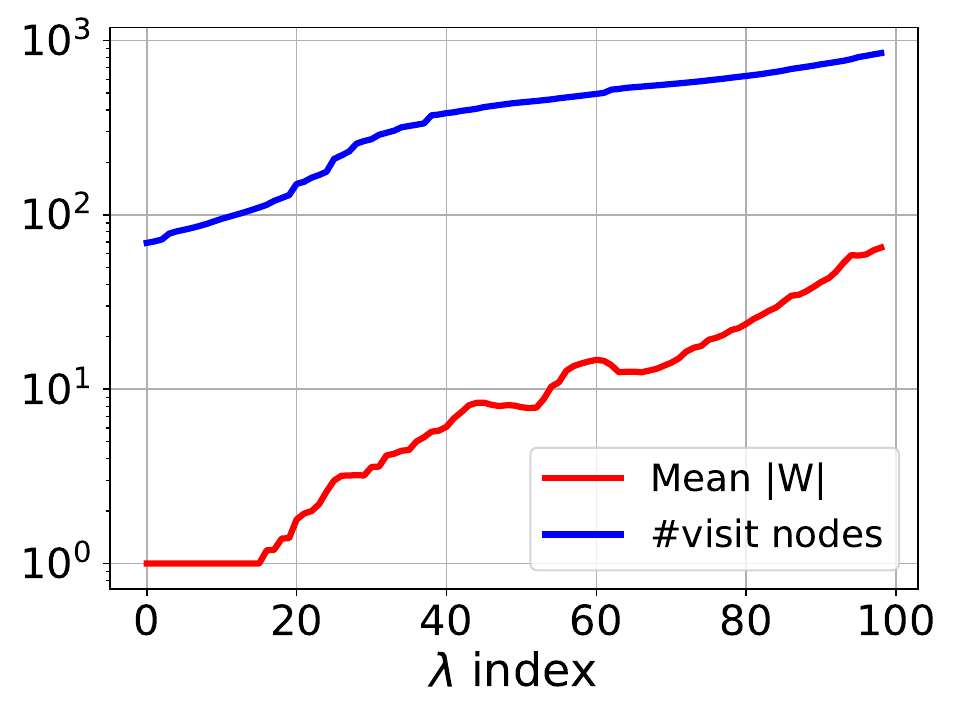}}
 \end{center}
 \caption{The average number of visited tree nodes and $|\cW|$.}
 \label{fig:num-traversed}
\end{figure}

\figurename~\ref{fig:num-traversed} shows the average number of traversed graph mining tree nodes and the size of selected $|\cW|$ for AIDS and ENZYMES. 
In this figure, both the values increased with the decrease of $\lambda$ because the effect of the sparse penalty becomes weaker.
As shown in \tablename~\ref{tab:ws_max_size}, the total number of the tree nodes were \red{
$134281$
and more than $15464000$ for AIDS
and ENZYMES, respectively.
\figurename~\ref{fig:num-traversed} shows the number of traversed nodes were at most about 
$6 \times 10^3 / 134281 (\approx 0.05)$
and $9 \times 10^2 / 15464000 (\approx 6 \times 10^{-5})$, respectively. }
This clearly indicates that our pruning strategy can drastically reduce the tree nodes, at which the evaluation of $g_H(\*\beta)$ is required as shown in Algorithm~\ref{alg:gpruning}. 
In the figure, we can see that $|\cW|$ was further small. 
This indicates the updated $\*\beta$ was highly sparse, by which the update of $\*z_v^H$ becomes easier because it requires only for non-zero $\beta_H$.

\section{Conclusion}
\label{sec:conclusion}

This paper proposed LAGRA (Learning Attributed GRAphlets), which learns a prediction model that linearly combines attributed graphlets (AGs).
%
In LAGRA, graph structures of AGs are generated through a graph mining algorithm, and attribute vectors are optimized as a continuous trainable parameters.
To identify a small number of AGs, the $L1$ sparse penalty is imposed on coefficients of AGs.
We employed a block coordinate update based optimization algorithm, in which an efficient pruning strategy was proposed by combining the proximal gradient update and the graph mining tree search.
%
%
Our empirical evaluation showed that LAGRA has superior or comparable performance with standard graph classification algorithms.
We further demonstrated that LAGRA actually can identify a small number of discriminative AGs that have high interpretability.


\section*{Acknowledgments}

This work was partially supported by MEXT KAKENHI (21H03498, 22H00300, 23K17817), and International Joint Usage/Research Project with ICR, Kyoto University (2023-34).

\clearpage

\bibliographystyle{IEEEtran}
\bibliography{ref}

\clearpage

\noindent
{\LARGE {\bf Supplementary Appendix for \\
 ``Learning Attributed Graphlets: Predictive Graph Mining by Graphlets with Trainable Attribute''}}

\appendix

\section{Update $\beta_0$}
\label{app:update-beta0}

The objective of $\beta_0$ can be re-written as
\begin{align}
      \min_{\beta_0} \ \sum_{i = 0}^n \max(1 - y_i (\*\psi_i^\top \*\beta + \beta_0), 0)^2 = \min_{\beta_0} \ \sum_{i \in \cI} (1 - y_i (\*\psi_i^\top \*\beta + \beta_0))^2.
      \label{eq:update-beta0-appendix}
\end{align}
For simplicity, we assume that all 
$\*\psi_i^\top \*\beta$ 
for 
$i \in [n]$ 
have different values (even when this does not hold, the optimal solution can be obtained by the same approach).
Depending on $\beta^{\rm (new)}_0$, elements in 
$\cI^{\rm (new)} = \{i \mid 1 - y_i(\*\psi_i^\top \*\beta + \beta^{\rm (new)}_0) > 0\}$
changes in a piecewise constant manner.
The point that $\cI^{\rm (new)}$ changes are characterized by the solution of the equation
$y_i (\*\psi_i^\top \*\beta + \beta_0) = 1 (i \in [n])$ with respect to $\beta_0$, 
i.e., there exist $n+1$ segments on the space of $\beta_0 \in \RR$.
Let $B^{(k)} = [\beta^{s_k}_0,\beta^{e_k}_0]$ be the $k$-th segment ($k \in \{0,\ldots,n\}$) and $\cI^{(k)}$ is $\cI^{\rm (new)}$ when $\beta^{\rm (new)}_0 \in B^{(k)}$.
Note that 
$\beta^{e_k}_0 = \beta^{s_{k+1}}_0 (k \in [n])$, which is the solution of 
$y_k (\*\psi_k^\top \*\beta + \beta_0) = 1$, and $\beta^{s_0}_0 = - \infty$ and $\beta^{e_n}_0 = \infty$.
Under the assumption of $\beta_0 \in B^{(k)}$, the optimal $\beta_0$ is
\begin{equation*}
 \hat\beta^{(k)}_0 =
  \frac{\sum_{i \in \cI^{(k)}}(y_i - \*\psi_i^\top \*\beta) }{ | \cI^{(k)} | }.
\end{equation*}
Since \eq{eq:update-beta0-appendix} is convex with respect to $\beta_0$, the obtained $\hat\beta^{(k)}_0$ must be the optimal solution if it satisfies $\hat\beta^{(k)}_0 \in B^{(k)}$.
Thus, the optimal $\beta_0$ can be found by calculating $\hat\beta_0^{(k)}$ for all $k \in \{ 0, \ldots, n \}$.

\section{Proof of Theorem~2.1}
\label{app:proof-theorem2-1}

From the definition of AGIS, the following monotonicity property is guaranteed:
\begin{align*}
 L(H^\prime) \sqsupseteq L(H) 
 \text{ and }
 H,H^\prime \in \overline{\cW}
 \ \Rightarrow \ 
 \psi(G_i; H) \geq \psi(G_i; H^\prime) 
\end{align*}
Let $M(G_i ; H)$ be the set of injections $M$ between $G_i$ and $H$.
The above monotonicity property can be easily verified from the fact 
\begin{align*}
 \min_{m \in M(G_i ; H)} D_{H,G_i}^{(m)} \leq \min_{m\in M(G_i ; H^\prime)}D_{H^\prime,G_i}^{(m)}. 
\end{align*}

Define
\begin{equation*}
      \begin{cases}
            a_i = y_i (1 - y_i (\*\psi_i^\top \*\beta + \beta_0)) & \mathrm{if}  ~~{ i \in \cI}, \\
            a_i = 0                                               & \mathrm{otherwise}.
      \end{cases}
\end{equation*}
Note that the sign of $a_i$ is same as $y_i$.
Using $a_i$, we re-write $g_{H^\prime}(\bm\beta)$ as
\begin{align*}
 g_{H^\prime}(\bm\beta)
 & = \sum_{ i \in \cI} a_i \psi(G_i; H^\prime) \\
 & = \sum_{ i \in \cI \cap \{ i \mid y_i > 0 \}}
      a_i \psi(G_i; H^\prime)
      + \sum_{ i \in \cI \cap \{ i \mid y_i < 0 \}}
      a_i \psi(G_i; H^\prime).
\end{align*}
From the monotonicity inequality
$\psi(G_i; H) \geq \psi(G_i; H^\prime)$, 
we see
\begin{align*}
 & \sum_{ i \in \cI \cap \{ i \mid y_i < 0 \}} a_i \psi(G_i; H)
 \leq
 \sum_{ i \in \cI \cap \{ i \mid y_i < 0 \}} a_i \psi(G_i; H^\prime)
 \leq
 g_{H^\prime}(\bm\beta)
 \\
 & \qquad
 \leq
 \sum_{ i \in \cI \cap \{ i \mid y_i > 0 \}} a_i \psi(G_i; H^\prime)
 \leq
 \sum_{ i \in \cI \cap \{ i \mid y_i > 0 \}} a_i \psi(G_i; H).
\end{align*}
From these inequalities, we obtain
\begin{equation*}
      \vert g_{H^\prime}(\bm\beta)\vert
      \leq
      \max\left\{
      \sum_{ i \in \cI \cap \{ i \mid y_i > 0 \}} a_i \psi(G_i; H),
      - \sum_{ i \in \cI \cap \{ i \mid y_i < 0 \}} a_i \psi(G_i; H)
      \right\}.
\end{equation*}

\section{Derivation of $\lambda_{\max}$}
\label{app:lambda_max}

When
$\*\beta = \*0$, 
the objective function of $\beta_0$ is written as the following piecewise quadratic function:
\begin{equation*}
      \min_{\beta_0} \frac{1}{2}\sum_{i \in \cI(\beta_0)} (1-y_i\beta_0)^2~~\mathrm{s.t.}
       \ \cI(\beta_0) =
      \begin{cases}
       \{ i \mid y_i > 0\}
       & \beta_0 \leq -1,    \\
       [n]
       & \beta_0 \in [-1,1], \\
       \{ i \mid y_i < 0\}
       & \beta_0 \geq 1.
      \end{cases}
\end{equation*}
%
In the region $\beta_0 \leq -1$, the minimum value is achieved by $\beta_0 = -1$, and for $\beta_0 \geq 1$, the minimum value is achieved by $\beta_0 = 1$. 
This indicates that the optimal solution should exist in $\beta_0 \in [-1,1]$ because the objective function is a smooth convex function. 
Therefore, the minimum value in the region $\beta_0 \in [-1,1]$ achieved by $\beta_0 = \sum_{i \in [n]} y_i / n$, defined as $\bar{y}$, becomes the optimal solution.

When $\*\beta = \*0$ and $\beta_0 = \bar{y}$, we obtain
\begin{equation*}
 g_H(\*0) = \sum_{i\in[n]} y_i \psi(G_i ; H) (1 - y_i\bar{y}).
\end{equation*}
From \eq{eq:gradient-condition}, we see that $| g_H(\*0) | = \lambda$ is the threshold that $\beta_H$ have a non-zero value.
This means that $H$ having the maximum $|g_H(\*0)|$ is the first $H$ that start having a non-zero value by decreasing $\lambda$ from $\infty$.
Therefore, we obtain
\begin{equation*}
      \lambda_{\max} = \max_{H\in\cH}
      \left\vert
      \sum_{i\in[n]} y_i \psi(G_i ; H) (1 - y_i\bar{y})
      \right\vert.
\end{equation*}

\section{Statistics of datasets}
\label{app:statistics-datasets}

Statistics of datasets is show in \tablename~\ref{tab:datasets}.

\begin{table*}
 \centering
 \caption{
 Statistics of datasets.
 }
 \label{tab:datasets}
 \begin{tabular}{|l|c|c|c|c|c|c|c|}
  \hline
 		 & AIDS   & BZR    & COX2   & DHFR   & ENZYMES      & PROTEINS     & SYNTHETIC   \\
  \hline
\# instances & 2000 & 405 & 467 & 756 & 200 & 1113 & 300 \\ 
Dim. of attribute vector $d$ & 4 & 3 & 3 & 3 & 18 & 1 & 1 \\
Avg. \# nodes & 15.69 & 35.75 & 41.22 & 42.43 & 32.58 & 39.06 & 100.00 \\
Avg. \# edges & 16.20 & 38.36 & 43.45 & 44.54 & 60.78 & 72.82 & 196.00 \\
\hline
 \end{tabular}
\end{table*}

\section{Additional Examples of Selected AGs}
\label{app:}

\begin{figure*}
 \begin{center}  
  \subfloat[]{\igr{.4}{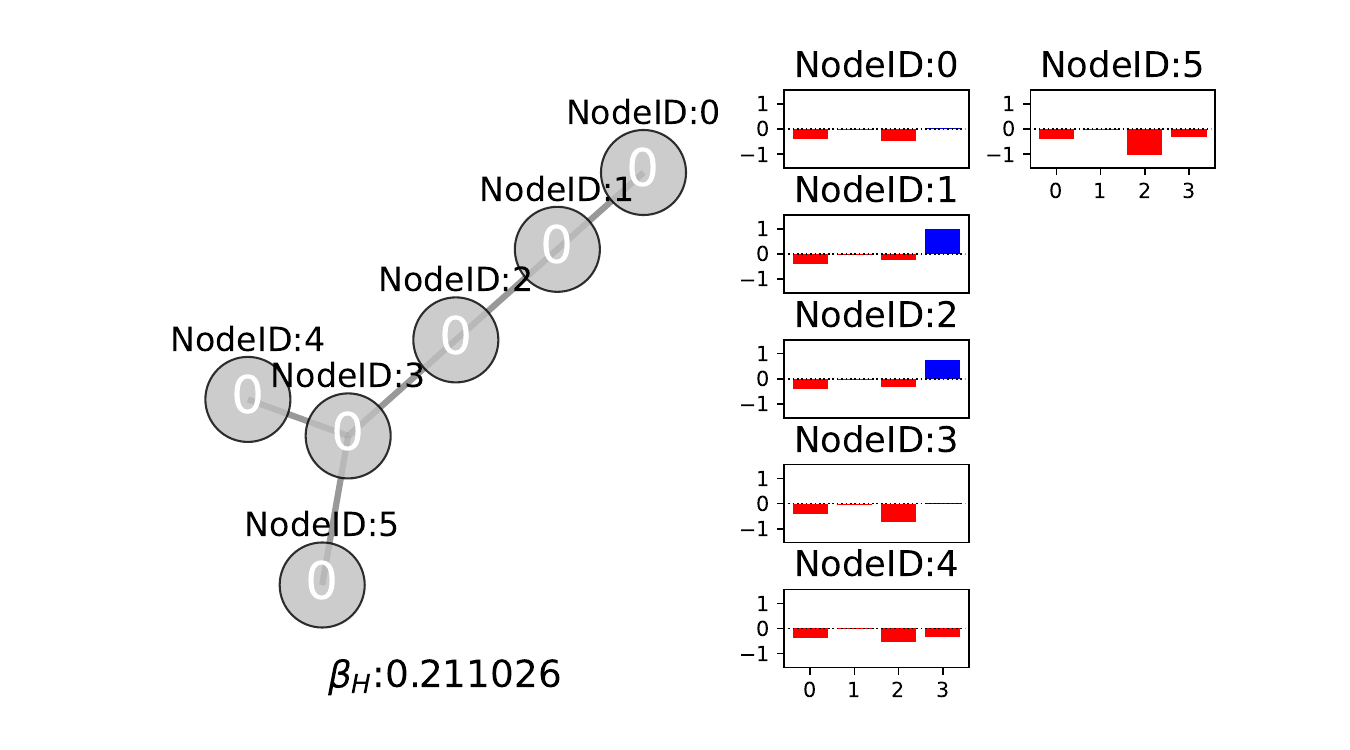}}
 \subfloat[]{\igr{.4}{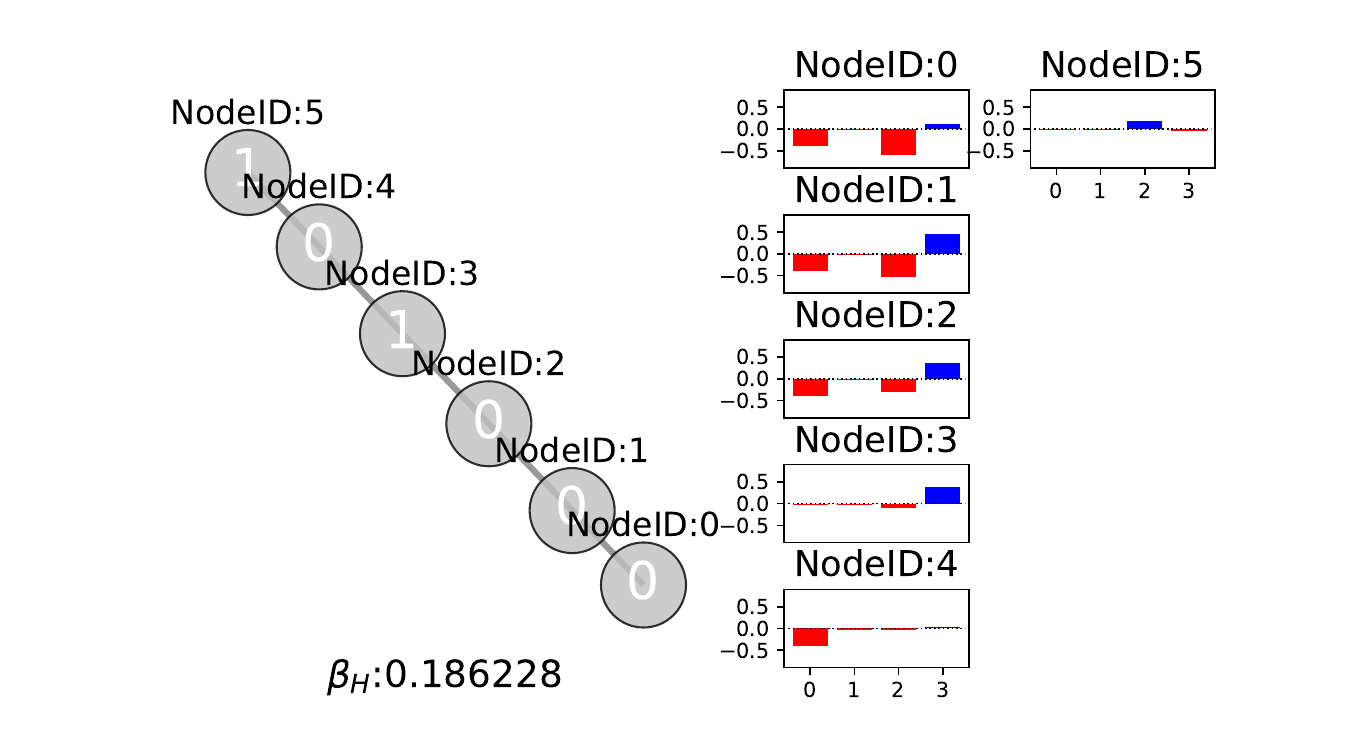}}

  \vspace{-1em}

  \subfloat[]{\igr{.35}{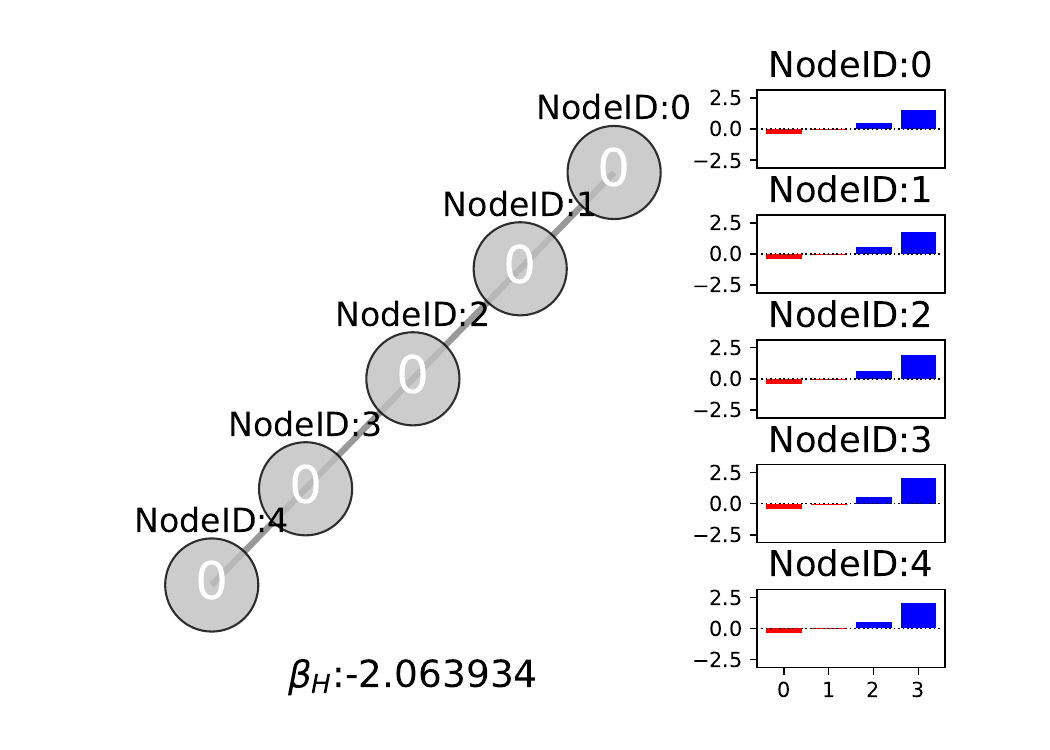}}
  \subfloat[]{\igr{.4}{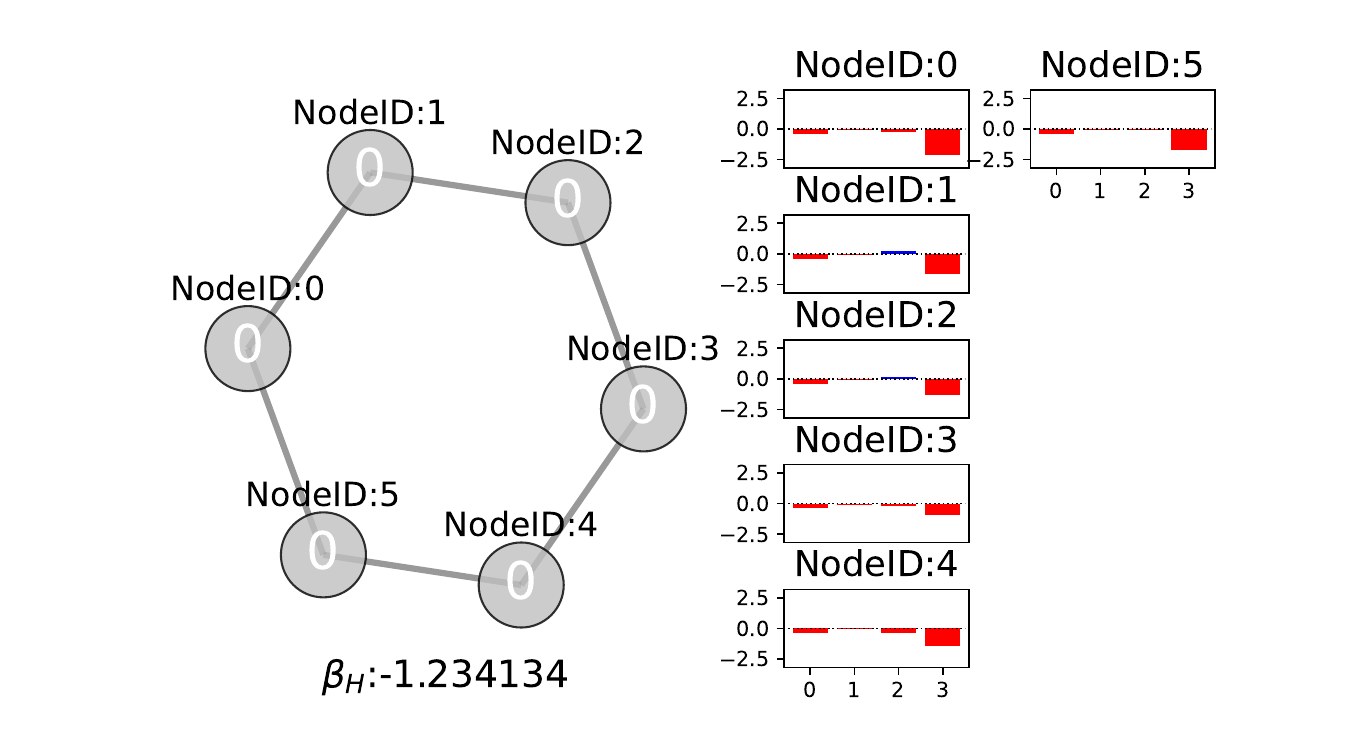}}
 \end{center}
 \caption{
 Selected important AGs for AIDS dataset.
 (a) and (b): AGs for the two largest positive coefficients. 
 (c) and (d): AGs for the two largest negative coefficients. 
 }
 \label{fig:AGs-AIDS}
\end{figure*}

\figurename~\ref{fig:AGs-AIDS} shows selected AGs for the AIDS dataset.

\end{document}